# Can Evidence Be Combined in the Dempster-Shafer Theory

John Yen
USC / Information Sciences Institute
4676 Admiralty Way
Marina del Rey, CA 90292

## Abstract

Dempster's rule of combination has been the most controversial part of the Dempster-Shafer (D-S) theory. In particular, Zadeh has reached a conjecture on the noncombinability of evidence from a relational model of the D-S theory. In this paper, we will describe another relational model where D-S masses are represented as conditional granular distributions. By comparing it with Zadeh's relational model, we will show how Zadeh's conjecture on combinability does not affect the applicability of Dempster's rule in our model.

## 1. Introduction

Zadeh has suggested a relational model for the Dempster-Shafer (D-S) theory [Zadeh, 1984a,b]. The relational model provides a clear and simple picture of the D-S theory. From this model, Zadeh has reached a conjecture on the combinability of evidence in the D-S theory [Zadeh, 1986a] [Zadeh and Ralescu, 1986b]. Since the Dempster-Shafer theory has attracted much attention in AI community recently, it is important to clarify several issues related to Zadeh's conjecture.

In the next section, we review Zadeh's relational model and his conjecture on the combinability of evidence. Section three describes our relational model for the D-S theory and discusses the combinability of our masses. Finally, we compare the two models and interpret Zadeh's conjecture using probabilistic terms.

## 2. Zadeh's Relational Model

The basic idea behind Zadeh's relational models of the D-S theory is to represent the mass distribution as a *granular distribution*, which is the set values distribution of a relation's attribute. For example, consider the following employee relation EMP in [Zadeh, 1986a]:

---

This article is based on the author's Ph.D. thesis at the University of California, Berkeley, which was supported by National Science Foundation Grant DCR-8513139.

| EMP | Name | Age |
|-----|------|---------|
|     | 1    | [22, 26] |
|     | 2    | [20, 22] |
|     | 3    | [30, 35] |
|     | 4    | [20, 22] |
|     | 5    | [28, 30] |

A granular distribution of employee's age can be obtained by summarizing the values of Age attributes in the reltion. Thus, we have

$$\Delta = \{([22,26], 1/5), ([20,22], 2/5), ([30,35], 1/5), ([28,30], 1/5)\} .$$

If the attribute's values are all singletons, its granular distribution corresponds to the probability distribution of the age of a randomly chosen employee.

The underlying relation EMP is called the *parent relation* of the granular distribution $\Delta$. A granular distribution has infinitely many parent relations because $\Delta$ is invariant under proportionally expanding the relation and permuting the values of Name.

As shown in [Zadeh, 1986a], D-S belief and plausibility measures correspond to the lower and upper bounds, respectively, on the fraction of employees who are within certain age range according to the relational database. Moreover, two granular distributions about an attribute obtained from two sources are combined in a way analogous to Dempster's combining rule.

### 2.1. The Combinability Problem

The combinability problem in Zadeh's relational model is based on two principles: (1) If an attribute is not allowed to take null values, two relations can not be combined if the combination results in null values. (2) Since granular distributions are summaries of relations, they can be combined only if at least a pair of their parent relations can be combined. This leads to Zadeh's conjecture that two granular distributions are not combinable if they do not have a conflict-free parent relation (i.e., a combined parent relation that does not have null values).

The conjecture significantly limits the applicability of Dempster's rule in Zadeh's model. In particular, two granular distributions cannot be combined if one contains a focal element that is disjoint from all focal elements of the other distribution. An even stronger condition has also been presented in [Zadeh and Ralescu, 1986].

## 3. Our Relational Model

In our relational model of the D-S Theory, granular distributions are explicitly conditioned on their background evidential sources. In order to express "conditioning" in the relational model, the sources are represented as attributes just like the frame of discernment is. In Zadeh's model, only the latter is represented as an attribute.

Suppose we know the sex distribution of employees in accouting department and a set-valued mapping from their sex to their age, the model can induce the age distribution of employees in accounting department just like the D-S theory does. Let us assume the sex distribution of employees in accounting department, deonted as a *conditional granular distribution*, is

$$\Delta_{Sex|Dept=Acct} = \{ (M, 1/4), (F, 3/4) \}.$$

A corresponding parent relation is tabulated below.

| EMP | Name | Age | Sex | Dept |
|---|---|---|---|---|
| | 1 | | F | Acct |
| | 2 | | M | Acct |
| | 3 | | F | Acct |
| | 4 | | F | Acct |
| | ... | | ... | Eng |

Given the following multivalued mapping from employees' sex to their age, one can induce the values of the later from the values of the former.

$\Gamma M=[20,22]$   $\Gamma F=[21,23]$

Hence, employees' ages in the relation are filled as shown below.

| EMP | Name | Age | Sex | Dept |
|---|---|---|---|---|
| | 1 | [21, 23] | F | Acct |
| | 2 | [20, 22] | M | Acct |
| | 3 | [21, 23] | F | Acct |
| | 4 | [21, 23] | F | Acct |
| | ... | ... | | ... |

From the relation, we can get a summary of age distribution for all the employees in accounting department:

$$\Delta_{Age|Dept=Acct} = \{([20,22],1/4) ([21,23],3/4)\}$$

Hence, the granular distribution of age in accounting department is determined by (1) the distribution of sex in the department and (2) the multivalued mapping from employees' sex to employees' age. This successfully models the way D-S masses are computed.

72

## 3.1. Combination of Evidence

Combination of conditional granular distributions differs from combination of Zadeh's granular distributions in that it is the **partial summaries** of an attribute, not the complete summaries, that gets combined. Combining two granular distributions that are conditioned on two pieces of evidence yields granular distribution that is conditioned on both of them. For instance, combining $\Delta_{Age|Dept=Acct}$ and $\Delta_{Age|State=CA}$ yields $\Delta_{Age|Dept=Acct, State=CA}$. Since the parent relations of conditional granular distributions are partially filled, there is more freedom in combining them. As a consequence, the conditional granular distributions are always combinable, as shown in the next section, as long as Dempster's rule is applicable.

## 3.2. Combinability of Conditional Granular Distributions

In our relational model, two conditional granular distributions have at least one conflict-free parent relation unless all the focal elements of one distribution are disjoint from those of the other. If the two conditional granular distributions does not have non-empty intersections, it is impossible to construct a conflict-free parent relation for the combined conditional granular distribution.

**Theorem:** Let $\Delta_{Age|E1}$ and $\Delta_{Age|E2}$ be two conditional granular distributions in the form of
$$\Delta_{Age|E1} = \{ (A_1, a_1/N), ..., (A_k, a_k/N) \}$$
$$\Delta_{Age|E2} = \{ (B_1, b_1/M), ..., (B_l, b_l/M) \}$$
where $\sum_{i=1}^{k} a_i = N$ and $\sum_{j=1}^{l} b_j = M$.

If their focal elements have at least one non-empty intersection, then there is at least one conflict-free parent relation for the combined distribution $\Delta_{Age|E1,E2}$.

**Proof:** Without loss of generality, let us assume $A_i$ and $B_j$ are two focal elements that have non-empty intersection. Also, let $\alpha = \min \{ a_i, b_j \}$. It is then straight forward to construct the conflict-free parent relation that is tabulated below.

| EMP | Name | Age1 | Age2 | E1 | E2 |
|---|---|---|---|---|---|
| | 1 | $A_i$ | $B_j$ | e1 | e2 |
| | ... | $A_i$ | $B_j$ | e1 | e2 |
| | $\alpha$ | $A_i$ | $B_j$ | e1 | e2 |
| | $\alpha+1$ | $A_1$ | | e1 | |
| | ... | ... | | e1 | |
| | N | $A_k$ | | e1 | |
| | N+1 | | $B_1$ | | e2 |
| | ... | | ... | | e2 |
| | N+M-$\alpha$ | | $B_l$ | | e2 |
| | ... | | | | |

Hence, the theorem is proved.

Although the theorem does not state that conditional granular distributions are always combinable. it does assure that they are combinable whenever Dempster's rule is applicable. When two mass distributions do not have non-empty intersections, the two evidential sources are mutually exclusive, i.e., they can not be present in the same time. Therefore, the total-conflict distributions indicates inconsistency and should not be combined at all. Indeed, Dempster's rule would not combine the distributions either, for the normalization factor is zero.

### 3.3. Assumptins of Dempster's Rule

While Dempster's rule seems a natural choice for combining Zadeh's granular distributions, Dempster's rule in our model implies certain assumptions. In fact, there are so many possibilities in combining conditional granular distributions that one always need to make certain assumptions, e.g., conditional independence assumption, in order to combine them. The conditional independence assumptions of a modified Dempster rule has been discussed in [Yen, 1986]. The formulation and discussion of the assumptions employed in original Dempster's rule is beyond the scope of this paper.

## 4. Discussion

### 4.1. A Comparison of The Two Models

Zadeh's granular distributions model D-S masses as unconditioned probability masses, while our conditional granular distributions model D-S masses as conditioned probability masses. Thus, different probabilistic interpretations of the D-S masses lead to different conclusions about their combinabilities. To offer a better insight to Zadeh's conjecture on the combinability of unconditioned mass, we discuss what the conjecture means in probabilistic terms in the next section.

### 4.2. A Probabilistic Interpretation of Zadeh's Conjecture

In probability terms, Zadeh's conjecture can be paraphrased as followed:
*Two unconditioned mass distributions, which are sets of constraints for the underlying probability distributions, cannot be combined if there does not exist a probability distribution that satisfies both constraint sets.*

**Definition:** A probability distribution P of a probability space $\Theta$ satisfies the constraints imposed by a mass distribution m if for every subsets A of space $\Theta$,
$Bel(A) \leq P(A) \leq Pls(A)$
where Belief and Plausibility functions are obtained from the mass distribution m.

**Lemma:** Suppose we have two relations $R_A$ and $R_B$ with same attribute "a". $A_i$ and $B_i$ denote values of attribute "a" at entry i of relation $R_A$ and $R_B$ respectively, as shown below.

| $R_A$ | Index | a |
|---|---|---|
| | 1 | $A_1$ |
| | 2 | $A_2$ |
| | ... | ... |
| | N | $A_N$ |

| $R_B$ | Index | a |
|---|---|---|
| | 1 | $B_1$ |
| | 2 | $B_2$ |
| | ... | ... |
| | N | $B_N$ |

If $B_i \subset A_i$, i = 1, 2, ...., N, we have
$$Bel_A(D) \leq Bel_B(D) \leq Pls_B(D) \leq Pls_A(D)$$
for every subset D of $\Theta$.

**Proof:** For all those entries that $A_i \subset D$, we have $B_i \subset A_i \subset D$.
Since belief of D is the relative count of those entries whose values are subsets of D, it follows that
$$Bel_A(D) \leq Bel_B(D)$$
and
$$Bel_A(D^c) \leq Bel_B(D^c).$$
Thus, we have
$$Pls_B(D) \leq 1 - Bel_B(D^c) \leq 1 - Bel_A(D^c) \leq Pls_A(D)$$
The lemma is, thus, proved.

Since each entry of the intersection of two conflict-free parent relations is subsumed by their corresponding entries, it follows from the lemma that the granular distribution of the combination of two conflict-free parent relations is a probability distribution that satisfies both constraint sets. Thus, if there exists conflict-free parent relations for two mass distributions, there must be at least one probability distribution that satisfies both constraint sets. Conversely, if there does not exist a probability distribution that satisfies both mass distributions, the two mass distributions do not have conflict-free parent relation and, cannot be combined.

## 5. Conclusions

In the context of relational model, the combinability of mass distributions in Dempster-Shafer theory depends on their interpretations. If D-S masses are viewed as unconditioned probability masses, they can be combined only if they have a conflict-free parent relation. However, if they are viewed as conditional probability masses, they are always combinable when Dempster's rule is applicable. As a result, Zadeh's conjecture on the combinability does not affect the applicability of Dempster's rule in our relational model.

The relational models not only provide a simple view to the D-S theory but also suggest that if we do not explicitly express mass distributions as conditioned on the evidential sources, we are confronted with the combinability problem.